\documentclass[letterpaper, 10 pt, conference]{ieeeconf}  

\IEEEoverridecommandlockouts                              

\usepackage{graphics} 
\usepackage{epsfig} 
\usepackage{mathptmx} 
\usepackage{times} 
\usepackage{amsmath} 
\usepackage{amssymb}  
\usepackage{xcolor}

\usepackage{multirow}
\usepackage{booktabs}
\usepackage{hyperref}

\usepackage{comment}

\newcommand{\FixColor}{black}

\title{\LARGE \bf
	F1 Hand: A Versatile Fixed-Finger Gripper for \\ Delicate Teleoperation and Autonomous Grasping
}

\author{Guilherme Maeda$^{1}$, Naoki Fukaya$^{1}$, and Shin-ichi Maeda$^{*1}$
	\thanks{$^{1}$1 Preferred Networks Inc. 1-6-1 Otemachi, Chiyoda, Tokyo, Japan
		{\tt\small [gjmaeda, fukaya, ichi]@preferred.jp}}
	\thanks{$^{*}$ Corresponding author.}
}

\begin{document}

	\maketitle
	\thispagestyle{empty}
	\pagestyle{empty}

	\begin{abstract}
		Teleoperation is often limited by the ability of an operator to react and predict the behavior of the robot as it interacts with the environment. For example, to grasp small objects on a table, the teleoperator needs to predict the position of the fingertips before the fingers are closed to avoid them hitting the table. For that reason, we developed the F1 hand, a single-motor gripper that facilitates teleoperation with the use of a fixed finger. The hand is capable of grasping objects as thin as a paper clip, and as heavy and large as a cordless drill. The applicability of the hand can be expanded by replacing the fixed finger with different shapes. This flexibility makes the hand highly versatile while being easy and cheap to develop. However, due to the atypical asymmetric structure and actuation of the hand usual grasping strategies no longer apply. Thus, we propose a controller that approximates actuation symmetry by using the motion of the whole arm. The F1 hand and its controller are compared side-by-side with the original Toyota Human Support Robot (HSR) gripper in teleoperation using 22 objects from the YCB dataset in addition to small objects. The grasping time and peak contact forces could be decreased by 20\% and 70\%, respectively while increasing success rates by 5\%. Using an off-the-shelf grasp pose estimator for autonomous grasping, the system achieved similar success rates to the original HSR gripper, at the order of 80\%.
	\end{abstract}

	\section{INTRODUCTION}

	Suction cups, widely used in logistics, are robust and simple 
	but usually only suited for grasping objects from the top and restricted by the presence of a surface appropriate for a vacuum seal.
	Dexterous hands are capable of complex manipulations, but are also expensive and difficult to control, 
	requiring careful consideration of their purpose before their adoption \cite{grebensteinAntagonisticallyDrivenFinger2010, akkayaSolvingRubikCube2019}.

	For this reason, simple structures such as two-finger grippers tend to be widely used. However, unlike suction cups, grippers need to be properly aligned to the object to be grasped. 
	%
	Under teleoperation, this positioning can be stressful and tiring, particularly when done over long periods of time.
	For example, when grasping a small object such as a screw on a desk with an adaptive finger, the operator needs to precisely predict the position of the fingertip before the finger is closed to avoid hitting the desk while assuring a successful grasp. This procedure requires careful and difficult adjustment of the spatial position of the finger often without visual references.
	In fact, it has been shown that teleoperation can take time to perform even on simple movements such as the pick up of a straw \cite{li_teleop_2017}.
	
	\begin{figure}
		\centering
		\vspace{0.2cm}
		\includegraphics[width=0.9\linewidth]{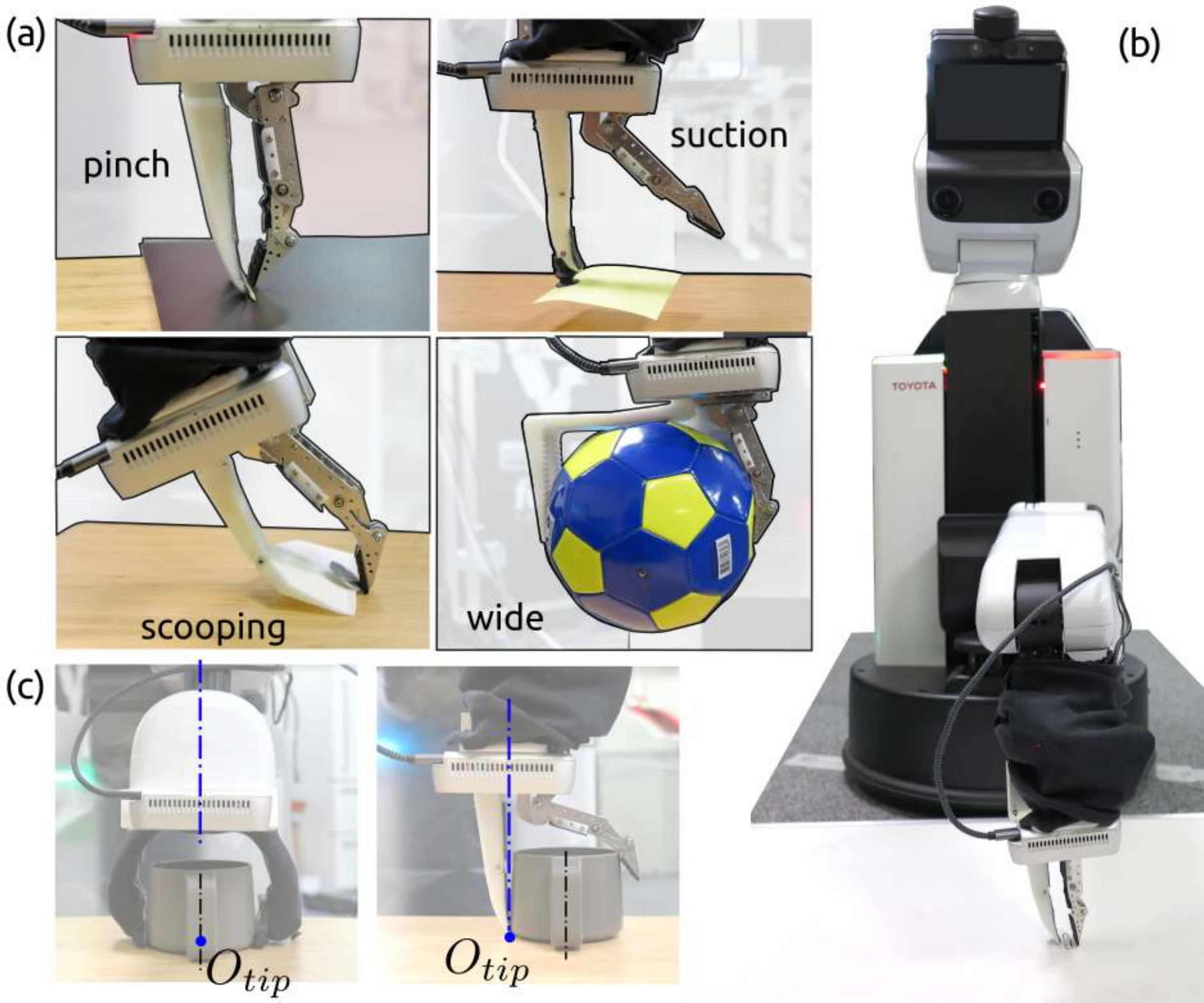}
		\vspace{-0.2cm}
		\caption{
				The F1 hand and some of its fingers.
				(a) The absence of actuation allows for easy shaping and switching of the fixed finger.
				(b) The F1 hand attached to a Toyota's HSR is capable of grasping objects as thin as a 0.7 mm thick clip.
				(c) We propose a control strategy to recover actuation symmetry as the atypical structure changes the origin of the gripper's tip reference frame  ($O_{tip}$) from the object center.
				Experiments can be watched by following the link: \href{https://youtu.be/iWXXIX4Mkl8}{https://youtu.be/iWXXIX4Mkl8}.
		}
		\label{fig:intro}
	\end{figure}
	To overcome such problems, we investigate a two-finger gripper with only one actuated finger, named F1 hand.
	By using the fixed finger as a reference, the operator can easily adjust the position of the fixed finger alongside the target object, which enables a more intuitive grasping in teleoperation.
	%
	{\color{\FixColor}
		Moreover, the adoption of a fixed finger not only decreases the cost of the hand, but it can be easily switched as the finger is free from actuators and electronic circuits. By tailoring the fixed finger it is possible to greatly expand the grasping capabilities of the F1 hand as shown in Fig. \ref{fig:intro}(a). 
	}

	However, as the closing motion of the fingers are asymmetrical, the control method for autonomous grasping with the F1 hand is not obvious, precluding one from using existing grasping methods for symmetrically actuated grasps. 
	As such, we propose a grasping strategy to enable the use of  existing grasping methods for symmetric grippers 
	by exploiting the entire arm motion during grasping {\color{\FixColor}(Fig. \ref{fig:intro}(c))}.
	
	For the evaluation of the resulting system, many studies evaluate the usefulness of the hand separately from the robot body \cite{grebensteinAntagonisticallyDrivenFinger2010, maM2GripperExtending2016,  achilliDesignSoftGrippers2020, KoTendonDrivenPFN} or tested with a prescribed motion of the robot arm in a predefined environment \cite{ bircherComplexManipulationSimple2021, yaguchiDevelopmentAutonomousTomato2016}. 
	However, in practice, it is not always possible to set the robot hand and body at a predefined position in a real environment. 
	To prove the practical usefulness of the F1 hand and its control methodology for autonomous grasping, we investigate the total grasping performance including the interaction between the robot’s motion and the gripper’s function.

	\section{RELATED WORK}
	There is a vast amount of research on robotic hands, with a wide variety of types and features.
	Even if we constrain the literature to only two-finger designs, the body of work can range from soft grippers with multi-segments \cite{teeple2020_hand} to designs specific for thin \cite{yoshimi2012_hand} and  small \cite{watanabe2021_hand} objects. In this section, we will analyze the characteristics of a typical two-fingered gripper and a gripper with one finger fixed.
	
	\subsection{Two-Finger Grippers}
	
	\begin{figure}
		\centering
		\vspace{.2cm}
		\includegraphics[width=0.9\linewidth, draft=false]{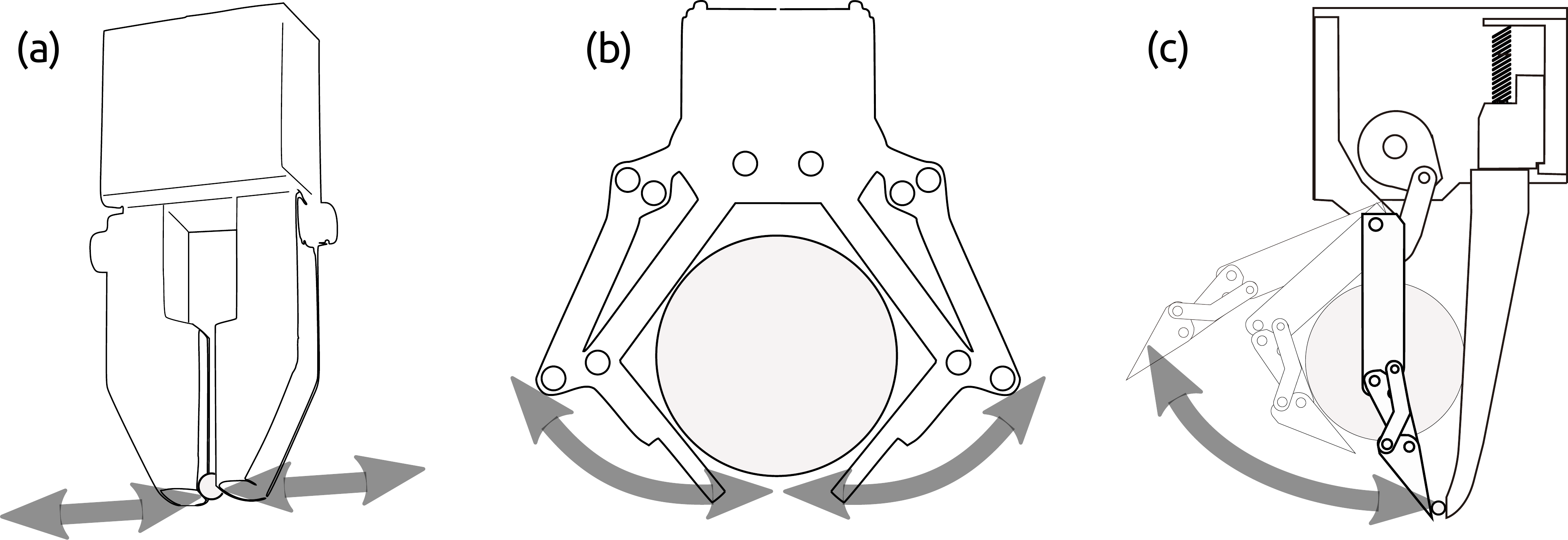}
		\caption{
			Three different types of two-finger gripper designs. From left to right, a parallel gripper, a multi-linkage adaptive finger, and our proposed F1 hand with a fixed finger.
			The arrows illustrate the motion of the tip of the finger.
		}
		\label{fig:gripper_type}
	\end{figure}
	
	As illustrated in Figs. \ref{fig:gripper_type}~(a,b), for our study, we categorize such grippers by the motion of the finger as grippers with parallel sliding fingers and multiple-linkage adaptive fingers.
	Parallel grippers are characterized by having the closure of the fingers done via linear motion and the grasping force is usually due to compression forces.
	Although some designs exist that eliminate this limitation (e.g.~\cite{kobayashiDesignDevelopmentCompactly2019}), parallel grippers usually have a limited maximum stroke as the amount of finger travel directly affects the size of the hand.
	For example, the Robotiq Hand-E has a stroke of 50 mm while the standard gripper of the Franka Emika has a stroke of 80 mm. 
	
\begin{figure}
		\centering
		\vspace{.2cm}
		\includegraphics[width=0.999\linewidth]{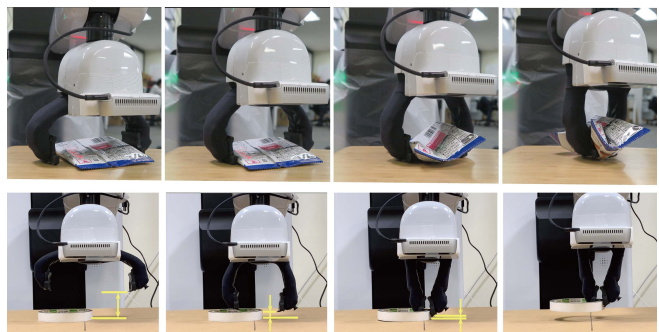}
		\caption{
			Top row: when grasping flat and wide objects on a hard surface, a teleoperator has to carefully move the hand upwards, as the fingers are closed.
			Bottom row: the hand must be placed at an appropriate height such that the finger does not hit the table surface before it makes contact with the object. 
		}
		\label{fig:hsr_f1_z_height}
	\end{figure}

	Multiple linkage fingers offer a better ratio between the size of the hand and the finger aperture.
	The multiple segments of the fingers are effective for grasping via robust mechanical enveloping (Fig. \ref{fig:gripper_type}(b)).
	Examples of such grippers are the Robotiq 2F-140, the Kinova gripper KG-2, the Toyota Human Support Robot (HSR) platform \cite{yamamoto2019development}, and our proposed gripper, presenting an opening distance between fingers of 140 mm,  175 mm, 125 mm, and 130.5 mm, respectively.
	Thus, a hand with a linkage mechanism can achieve a larger opening than a parallel gripper, but on the other hand, multiple linkage fingers make it difficult to intuitively know where the fingertip will be located when the finger is closed (Fig. \ref{fig:hsr_f1_z_height}).

	As a consequence, when performing top graspings by teleoperation, the fingertips may collide with the surface giving rise to unnecessary normal forces due to the difficulty in continuously adjusting the hand position or finding the appropriate position of the hand midair.
	{\color{\FixColor}
		Empirically, we found that when using the original HSR gripper, normal contact  forces beyond the range 50$\sim$60 N---sensed by the original HSR force-torque sensor mounted on the wrist---would at times raise an emergency stop signal, rendering the robot unresponsive.
		As such, within the scope of this paper, we consider delicate graspings to be those that generate  contact forces of less than 40 N.
	}
	
	\subsection{Fixed-Finger Grippers}
	
	\color{\FixColor}
	In general, a solution to decrease a mechanism's cost and increase its reliability is by simplifying the  design and reducing the number of parts required.
	For a two-finger gripper, this can be achieved by eliminating actuation in one of the fingers.
	While this loss of motion may require atypical grasping approaches, one upside is that switching and designing tailored fixed-fingers with different structures is much easier than for an actuated finger.
	\color{black}
	
	For example, ugo, a security robot, has this type of gripper. However, its grippers are mainly used to press elevator buttons and rarely to hold objects \footnote{Demonstration of ugo: {\color{\FixColor}\url{https://www.youtube.com/watch?v=9xybTXT9hYU}}}.
	Pastor et al. use this gripper to ensure a stable and large surface for mounting tactile sensors \cite{pastorUsing3dConvolutional2019}. The setup was used to investigate in-hand manipulation. In particular, palpatation movements due to the actuated finger were used to recognize objects using a deep neural network. 

	Inspired by surgical palpatation, the M$^2$ gripper \cite{maM2GripperExtending2016} is a gripper whose design is aimed at dexterous in-hand manipulation where the fixed finger acts as a stable thumb counter-part of the forefinger.
	To achieve dexterities such as in-hand controlled slip, two independent  actuators are required to allow for the control of the agonist and antagonist tendons of the forefinger.
	A similar design uses an active surface in the form of a belt on the fixed finger \cite{maInhandManipulationPrimitives2016}.
	In both cases, the grippers were evaluated with a static base---not as part of a grasping robot system---and at times with a fixed arm motion as their focus was on in-hand manipulation.
	The gripper was used in \cite{zhangDoraPickerAutonomousPicking2016} where experiments indicate that the capabilities of a simple vacuum gripper and a symmetrically actuated gripper are complimentary to the M$^2$ when grasping a wide range of objects.

	\section{The F1 Hand Design}
	
\begin{figure}
		\centering
		\vspace{0.3cm}
		\includegraphics[width=0.99\linewidth]{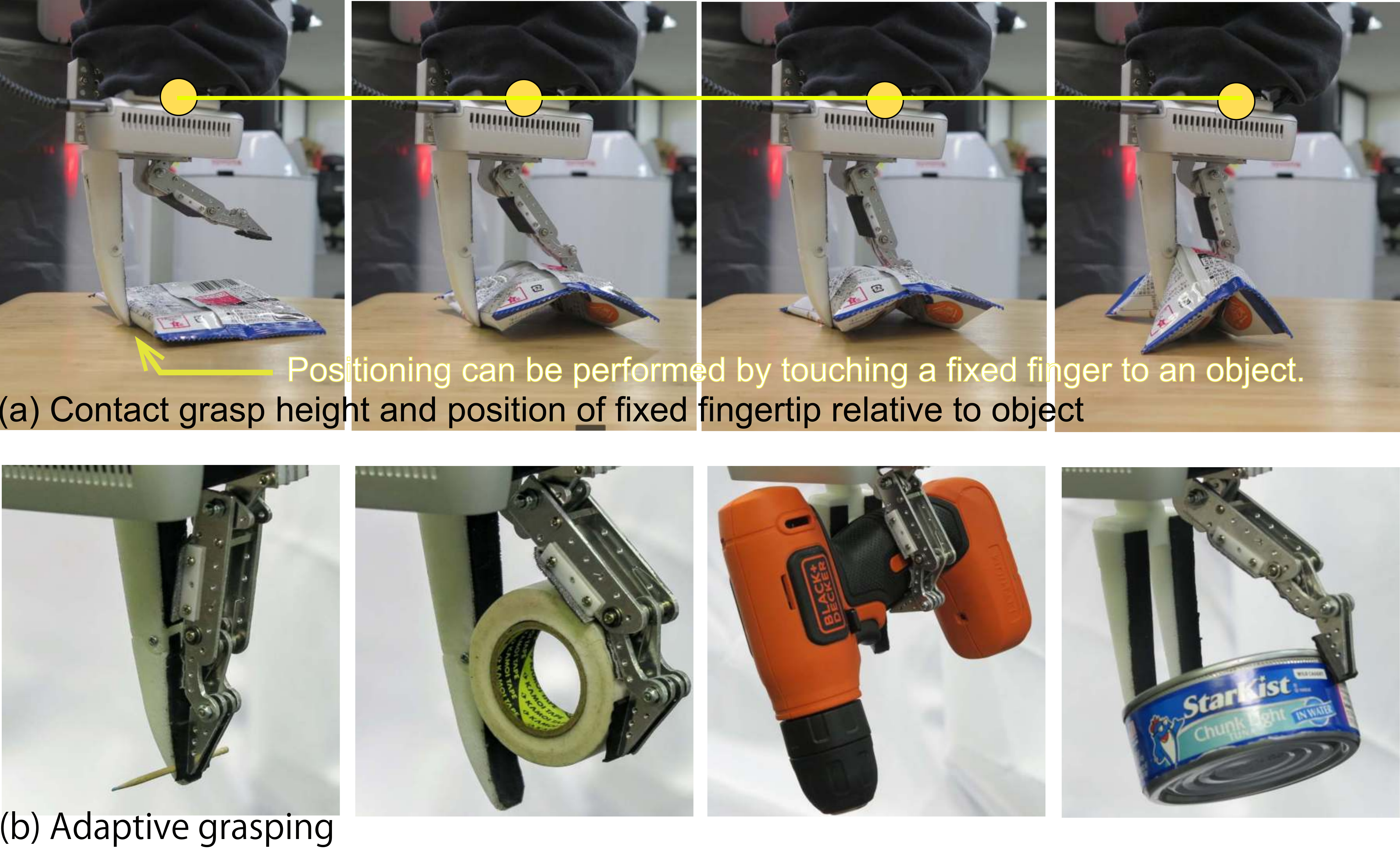}
		\caption{
			(a) The fixed finger provides a visual guide to check the distance to the object that also dictates a constant grasp height which is particularly useful in teleoperation.
			(b) The multi-linkage finger allows a continuum of adaptive grasps varying from pinch to enveloping grasps that only depends on the positions of the grasping interaction forces.
		}
		\label{fig:F1_advantages}
	\end{figure}

	The proposed F1 hand uses a fixed finger for three main purposes.
	First, the fixed finger provides a physical and fixed end-effector tip that dictates the constant grasp height between the object and the hand as shown in Fig. \ref{fig:F1_advantages}(a) and is particularly useful to facilitate delicate top grasp teleoperation.
	Second, as previously motivated, the ability to freely shape and change the configuration of the hand\footnote{A promotional video of the hand with different attachments can be watched here: \url{https://www.youtube.com/watch?v=sN66DilliBs}}.
	Third, as shown in Fig.\ref{fig:F1_advantages}(b) the fixed finger acts as the supporting counterpart of the multi-link actuated movable finger, allowing for adaptive enveloping or pinch grasps depending on the shape of the object.
	The hand combines the fixed-height grasping benefits of a parallel gripper with the advantages of the adaptive finger for delicate and robust grasps.
	
\begin{figure}
		\centering
		\includegraphics[width=0.85\linewidth]{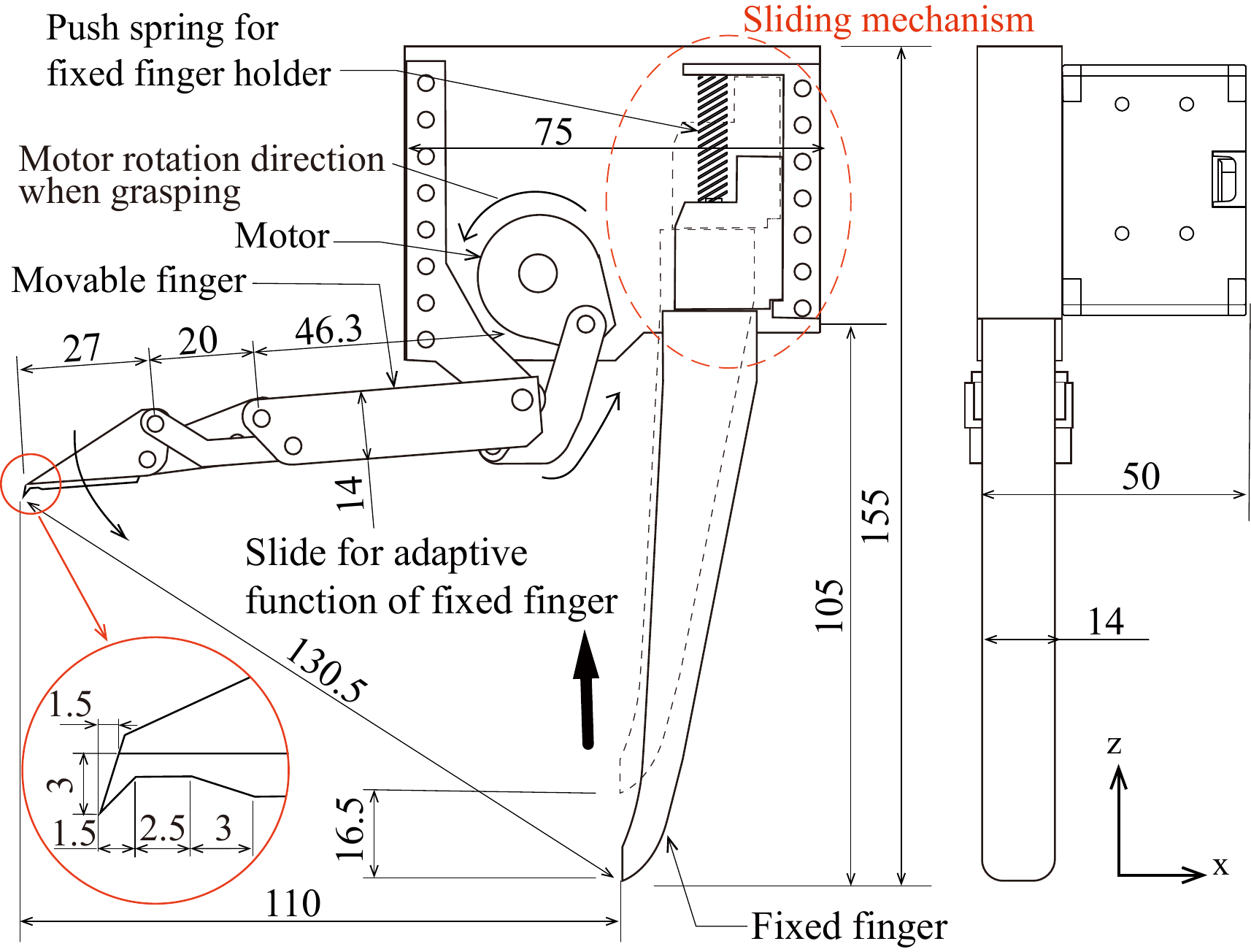}
		\caption{Structure {\color{\FixColor}of F1 hand having 	a fixed finger with} a sliding mechanism.}
		\label{fig:f1wholedesign}
	\end{figure}

	Fig.~\ref{fig:f1wholedesign} shows the structure and dimensions of the hand.
	The fixed finger is attached to a vertical linear sliding holder which is maintained in its initial extended state by a push spring.
	The slider allows the fixed finger to retract inside the palm under contact with objects or surfaces. 
	%
	%
	{\color{\FixColor}
		This retraction effectively creates a margin for positioning error.
		Not only the spring allows for the compensation of small errors in the approaching direction of the grasping, thus increasing success rates, but also protects the hand from damage.
	}

\begin{figure}
		\centering
		\vspace{0.2cm}
		\includegraphics[width=0.9\linewidth]{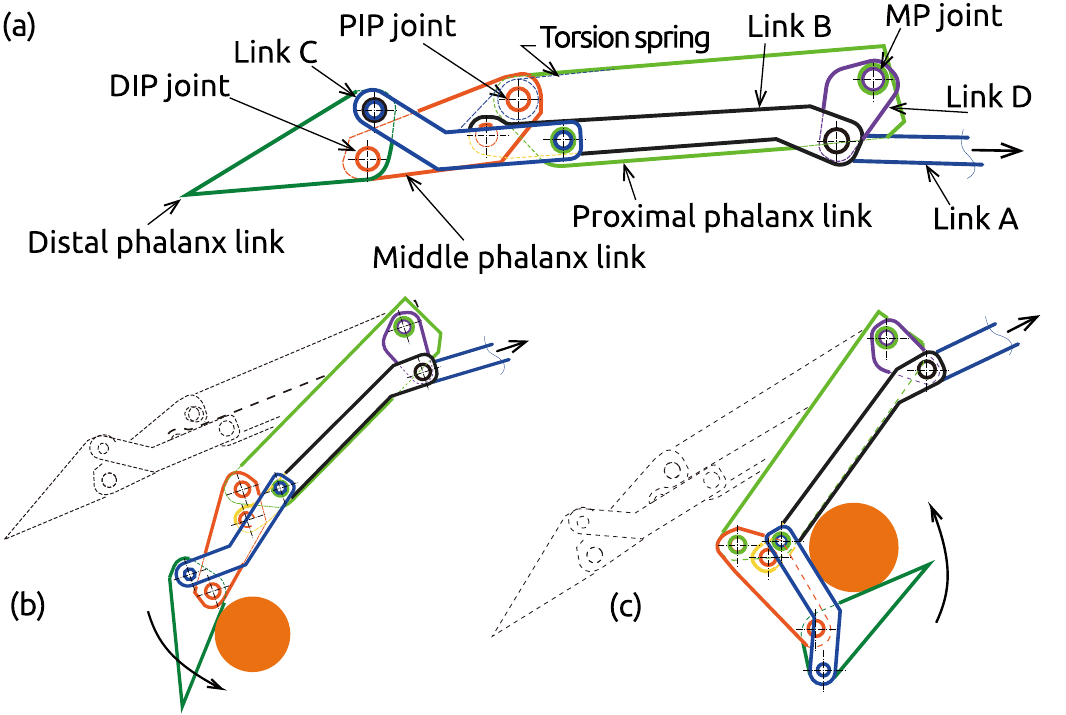}
		\vspace{-0.2cm}
		\caption{{\color{\FixColor}Illustration} of the design and motion of the movable finger.
			(a) Kinematic structure.
			(b) Final configuration under fingertip contact.
			(c) Final configuration under even contact where all three links envelop the object.
		}
		\label{fig:f1fingerdesign}
	\end{figure}

	As shown in Fig. \ref{fig:f1fingerdesign}(a), the actuated finger is a multiple linkage mechanism based on a hand developed by one of the authors in the past~\cite{fukaya2000_hand, fukaya2013_hand}, but here constructed compactly due to a modification of the link B. The size was determined based on the middle finger of
	{\color{\FixColor} an adult male}. The three joints (MP joint, PIP joint, and DIP joint) can be operated in conjunction by applying a driving force to one point.
	Two torsion springs are attached to the PIP joint of the movable finger to maintain the finger in extension as the basic posture, and the spring constant of the torsion spring is K=0.13 N$\cdot$mm/deg.
	
	The actuated finger is driven by a motor (Dynamixel XM430-W350-R). 
	The motor and the movable finger are connected by the Link A to transmit the driving force. 
	The driving state of the finger differs depending on which part of the finger contacts the object and allows for highly adaptive grasps.
	As shown in Fig.~\ref{fig:f1fingerdesign}(b), when the distal phalanx link contacts the object, the movable finger transmits the grasping force while maintaining the extended state.
	If the position of the object is shifted when the finger contacts the object, the finger automatically changes to a posture that allows stable contact.
	
	Conversely, as shown in Fig.~\ref{fig:f1fingerdesign}(c), if the proximal phalanx link touches the object, the link D moves independently and actuates the middle phalanx via link B, while the link C (which is connected to the proximal phalanx link) pulls the distal phalanx link. 
	As a result, the movable finger moves such that its segments wrap around the object, and the grasping motion automatically stops when all possible contact parts are touched. 
	As the contact area is maximized, the contact forces of each part is distributed and the object can be grasped stably even with small motor torques.

	\section{Grasping with the F1 Hand}
	
	When naively grasping small or thin objects with the F1 hand, one needs to carefully reason the reachability of the fingers.
	As shown in the {\color{\FixColor}left illustration} of Fig. \ref{fig:controlstrategy}(a), if the fixed finger is too far from the object, closing the gripper with the actuated finger while maintaining the hand static would eventually lead to the uncontrolled sliding or rolling of the object;
	{\color{\FixColor} which among other factors, depends on the unknown friction between the parts.}
	{\color{\FixColor}
		As a measure, one could position the fixed finger as close as possible to the object before executing the grasp as shown in the right case in Fig. \ref{fig:controlstrategy}(a). 
		However, during the pre-grasping motion, this poses the risk of the fixed finger accidentally hitting the object, disturbing it from its original position.
	}
	
	\subsection{Grasping in Teleoperation}    \label{sec:teleop_controller}
	
	\begin{figure}
		\centering
		\includegraphics[width=0.9997\linewidth]{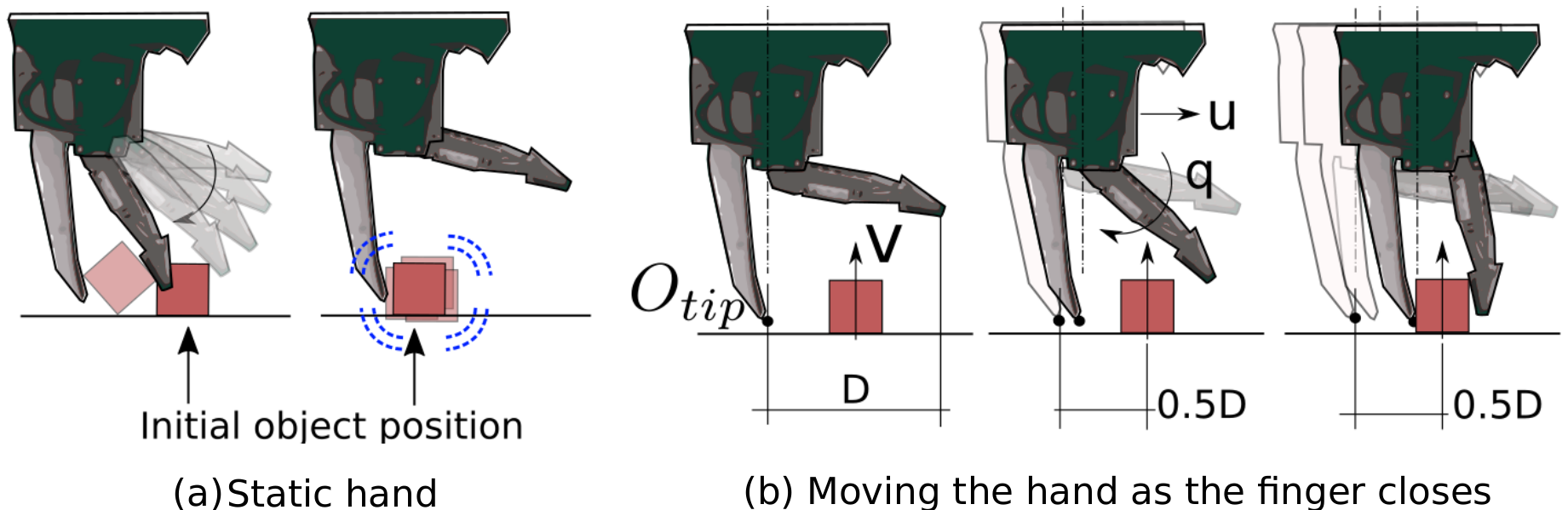}
		\caption{ {\color{\FixColor} (a) The fixed finger is too far from the object when the grasping starts and the actuated finger eventually rolls or slides the object (left), potentially leading to grasping failures.
				Starting the grasp close to the object can be difficult due to the required positioning requirements and possible disturbance on the initial object position (right).
				(b) A natural approach is to assume that the object lies somewhere in between the current aperture of the hand $D$.
		}}
		\label{fig:controlstrategy}
	\end{figure}
	Fig. \ref{fig:controlstrategy}(b) illustrates a strategy that resembles that of a conventional two-finger gripper.
	It assumes that the object is centered between the two fingers' projected distance on the grasping surface, indicated by the length $D$.
	To grasp, the object is approached by both fingers simultaneously. 
	This is done with the use of a grasping primitive that closes the actuated finger with a motor command $q$, while moving the entire hand (and thus robot) in the direction $u$ over a distance of $0.5 D$.
	
	The robot control is accomplished by moving the $O_{tip}$ horizontally as a Cartesian target towards the center of the object and solving the inverse kinematics for the whole robot motion.
	In the same control loop, incremental commands for the actuated finger are sent such that by the end of the process the actuated finger is closed.
	As the robot moves and the actuated finger closes, contact will be eventually made, resulting in the grasping of the object.
	If the object is perfectly centered, both fingers will meet the left and right walls of the object at the same time.
	%
	{\color{\FixColor}
		In practice, the level of accuracy in which an operator can center the object between the fingers highly depends on his/her experience with the system. However, in general, we noticed that the approach of using a grasping primitive of Fig. \ref{fig:controlstrategy}(b) is much more forgiving in terms of positioning error when compared to the strategies shown in Fig. \ref{fig:controlstrategy}(a).
	}
	
	\subsection{Autonomous Grasping} \label{sec:autonomous_grasping}
	
	\begin{figure}
		\centering
		\vspace{0.2cm}
		\includegraphics[width=0.95\linewidth]{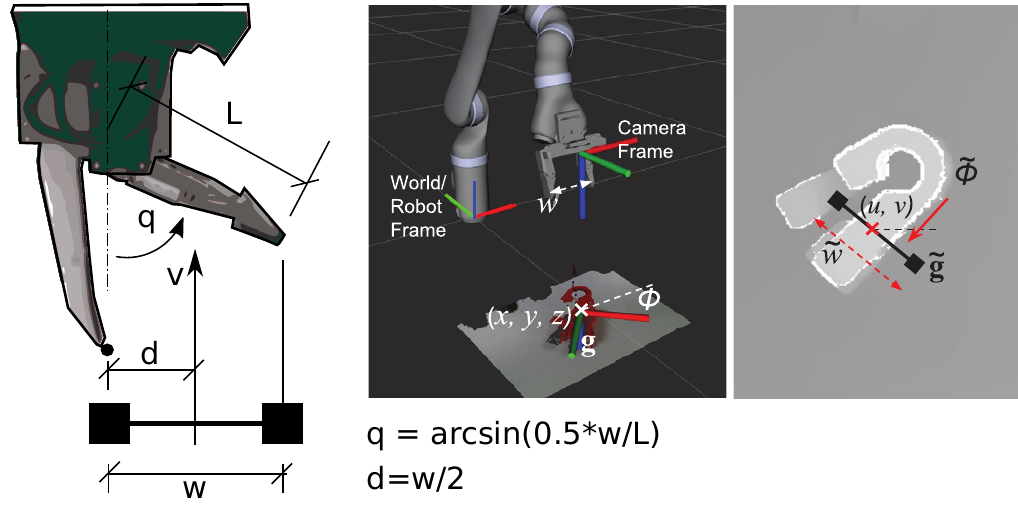}
		\vspace{-0.2cm}
		\caption{
			Parametrization of the grasping controller based on the values output by an off-the-shelf grasp pose estimator GG-CNN \cite{morrisonLearningRobustRealtime2020} (the right figure is adapted from the same paper).
		}
		\label{fig:controlstrategyposeestimator}
	\end{figure}
	
	\begin{figure}
		\centering
		\includegraphics[width=0.9\linewidth]{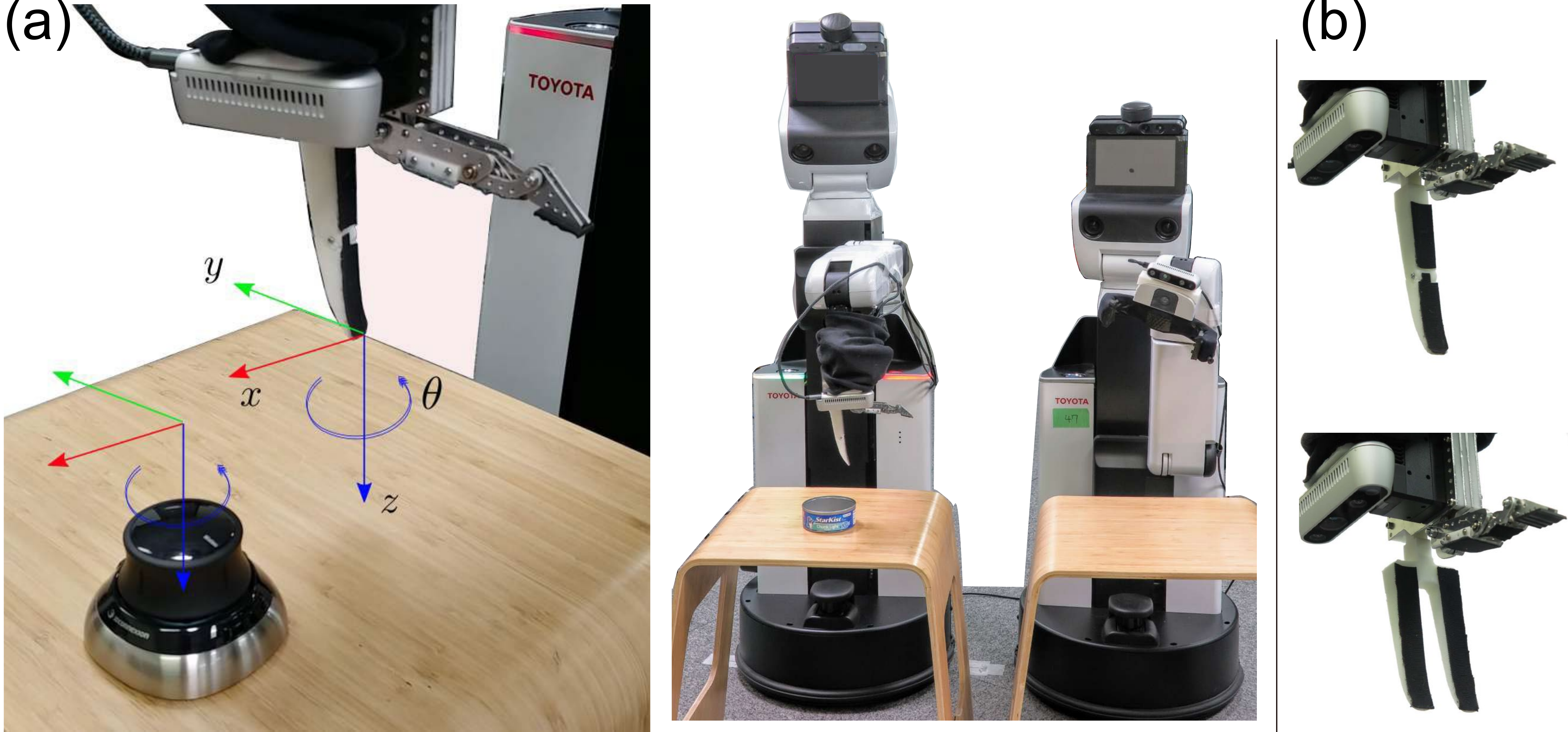}
		\vspace{-0.1cm}
		\caption{  
			Experimental setup. (a) The 3D mouse interface and {\color{\FixColor}the two HSRs used for experiments where the left HSR equipped with F1 hand shows the teleoperation initial position}. 
			(b) The thin fixed finger {\color{\FixColor}(upper)} was used for teleoperation experiments and the thicker version {\color{\FixColor}(lower)} was used for autonomous grasping experiments. 			
		}
		\label{fig:robot_setup}
	\end{figure}

	\begin{figure*}
		\centering
		\vspace{0.2cm}
		\includegraphics[width=0.95\linewidth]{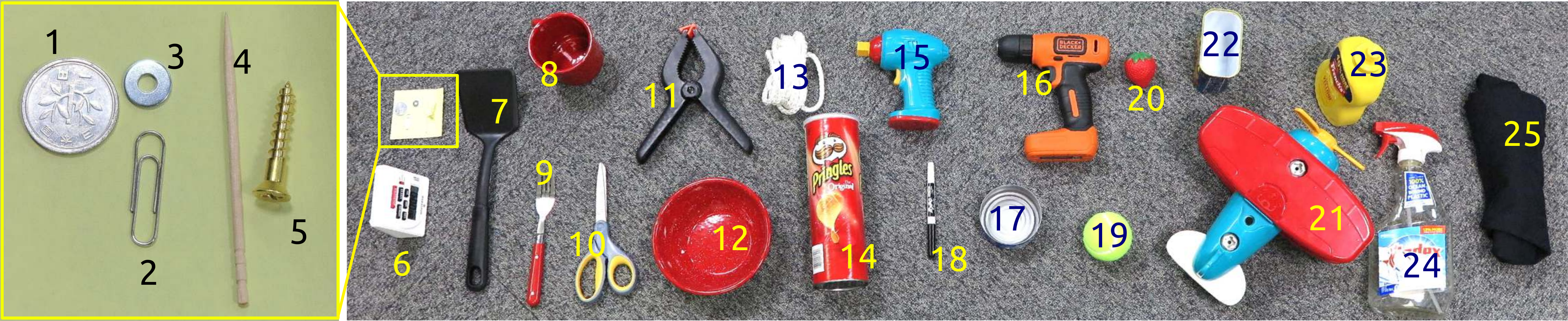}
		\vspace{-0.2cm}
		\caption{
			The objects used for experiments are based on a subset of YCB objects with the addition of the toothpick, the coin, and the paper clip.
			The pose of each object in the picture also indicates how they were placed on the table during top graspings.
			The numbers are used to identify each of the objects.
			Teleoperation experiments can be watched in the first part of the video here: \href{https://youtu.be/iWXXIX4Mkl8}{https://youtu.be/iWXXIX4Mkl8}
		}
		\label{fig:objects}
	\end{figure*}

	As shown in Fig. \ref{fig:controlstrategyposeestimator}, we assume that the robot is provided with an approaching direction as vector $v$ and that there are two points of contact (one for each finger) symmetrically opposed on a plane containing $v$ and separated by a distance $w$.
	This assumption is in accordance with many of the recent vision-based grasp pose estimators found in the literature, including 6-DoF estimators (e.g. \cite{mahlerDexnetDeepLearning2017, murali20196, morrisonLearningRobustRealtime2020, breyerVolumetricGraspingNetwork2021}).
	In the particular case of planar top grasps, $v$ is normal to the surface to which the camera point towards.
	As the $O_{tip}$ of the F1 hand is defined to be at the tip of the fixed finger we set its pre-grasping position to be $d=w/2$ and open the actuated finger to $w$.
	We then call the same primitive grasping controller used for teleoperation.

	The described process provides the parameters of the grasping primitive in which the input is the center of the object to be grasped.
	If a grasp pose estimator also provides the gripper aperture $w$, the value of $d$ and the initial angle of the actuated finger can be specified. 
	This allows us to treat the hand as if it was a parallel symmetrical gripper as shown in 
	Fig. \ref{fig:controlstrategyposeestimator} where the controller maps the output of the grasp pose estimator GG-CNN \cite{morrisonLearningRobustRealtime2020} to the asymmetric closing of the gripper.

	\section{Experiments} \label{sec:experiments}
	
	Two identical Toyota HSRs were used, one with the original gripper and another one with the F1 hand, as shown in Fig. \ref{fig:robot_setup}(a). 
	The F1 hand's motor, the 3D mouse, and the RealSense D435i were connected via USB cable. 
	The HSR was connected to a desktop PC via LAN and running ROS.
	In teleoperation, the operator sitting in front of the robot used a Connexion 3D mouse to move the reference position of the end-effector $O_{tip}$.
	To make a fair comparison with the original HSR gripper, we fixed the shape of the fixed finger to a narrow design (left picture in Fig. \ref{fig:robot_setup}(b)) and did not change it during the teleoperation experiments.
	{\color{\FixColor}
		The contact forces were measured using the original force-torque sensors of the HSR, located at the flange part of the gripper mount. 
		The F1 hand and its mounting point were designed such that the attachment of the F1 hand did not affect the functionality of the force-torque sensor.
	}
	
	In autonomous grasping, we implemented the  GG-CNN~\cite{morrisonLearningRobustRealtime2020} as a popular top grasp pose estimator and mapped its output parameters to the grasping controller.
	The GG-CNN used the RealSense D435i camera attached to the back of the hand of the robot.
	We used a wider version of the fixed finger to increase the robustness to inaccurate pose estimations and calibration errors (right finger in Fig. \ref{fig:robot_setup}(b)).
	The experiments can be watched by following the link: \href{https://youtu.be/iWXXIX4Mkl8}{https://youtu.be/iWXXIX4Mkl8}

	\subsection{Experiments in Teleoperation}

	As shown in Fig. \ref{fig:objects}, 22 objects from the YCB dataset  \cite{calliYaleCMUBerkeleyDatasetRobotic2017a} were selected and were added with three additional smaller and thinner objects: a Japanese one-Yen coin (diameter 20 mm, thickness 1.5 mm), a small clip (length 27 mm, thickness 0.7 mm), and a toothpick (length 60 mm, diameter 2 mm).
	Together with the YCB small washer (diameter 10 mm, thickness 1.2 mm), these additional set of objects become extremely challenging for general-purpose grippers.
	
	Although many authors have proposed grippers that can grasp small objects (e.g. \cite{yoshimi2012_hand, watanabe2021_hand}), such designs render the gripper  tailored to a specific range of sizes.
	Conversely, here we use the YCB subset in conjunction with our added small objects to provide evidence that our hand can cover a wide variety of shapes, sizes, and weights.
	The subset of 22 objects, shown in  Fig. \ref{fig:objects}, were selected based on their shape and weight.
	
	We favored objects with curvature, excluding the many package boxes, Lego blocks, and wooden blocks due to their trivial grasping geometry. 
	The mug (\#8) and the Pringles can (\#14) were selected as representative of the many cups and other cylindrical objects of equivalent size.
	Plastic containers with complex curvatures are represented by the mustard and the Windex bottle (\#23, \#24).
	The tennis ball (\#19) and the plastic strawberry (\#20) were selected to represent the numerous rounded plastic objects shaped as fruits.
	Among the many washers, screws, and nuts of the dataset, we selected a small washer of the dataset (\#3) and a screw (\#5).
	The electric drill (\#16) and the toy airplane (\#21) were selected due to their large size and weight.
	
	Our selection of objects covered the following ranges.
	The heaviest object was the hand drill (600 g). 
	The thinnest object was the paperclip  (0.7 mm).
	The smallest object was the washer ($\phi$10 mm $\times$ 1.2 mm)
	The biggest object was the toy airplane (bounding box of approximately 26 cm $\times$ 28 cm $\times$ 20 cm).
	Other notable objects that were not included in the experiments are discussed in Sec.  \ref{sec:limitation_and_extension}.
	
	As one of the most basic and frequently used motions in teleoperation, we limited the scope of experiments to top grasps only.
	As such, only the yaw angle $\theta$ was controlled together with the ($x,y,z$) directions (see Fig. \ref{fig:robot_setup}).
	%
	{\color{\FixColor}
		For each object, 20 to 21 teleoperation trials were executed by an expert operator whose goal was to grasp as fast as possible.
		To the best of our efforts, we believe the effects of learning curve and biases towards one of the gripper designs were not present. 
		The teleoperator was used to the 3D mouse interface and we assured he had been practicing graspings for a few weeks with both the original HSR gripper and the F1 hands before the experiments were recorded.
	}
	
	The trials were randomized by fixing the object and robot's initial poses while randomly setting the initial rotation of the wrist by sampling the rotation angle from a uniform distribution in the range $[-45^{\circ}, 45^{\circ}]$.
	The random initial rotation forced the operator to teleoperate the robot to bring the gripper to a pre-grasp position.
	Once the pre-grasping position was achieved, the operator pressed a button that in the case of the HSR gripper activated the torque-controlled grasp, and in the case of F1 hands executed the primitive described in {\color{\FixColor}Sec. \ref{sec:teleop_controller}}.
	The robot then lifted the object 10 cm such that the operator could visually confirm the grasp success.

\begin{figure}
		\centering
		\vspace{.2cm}
		\includegraphics[width=.94\linewidth]{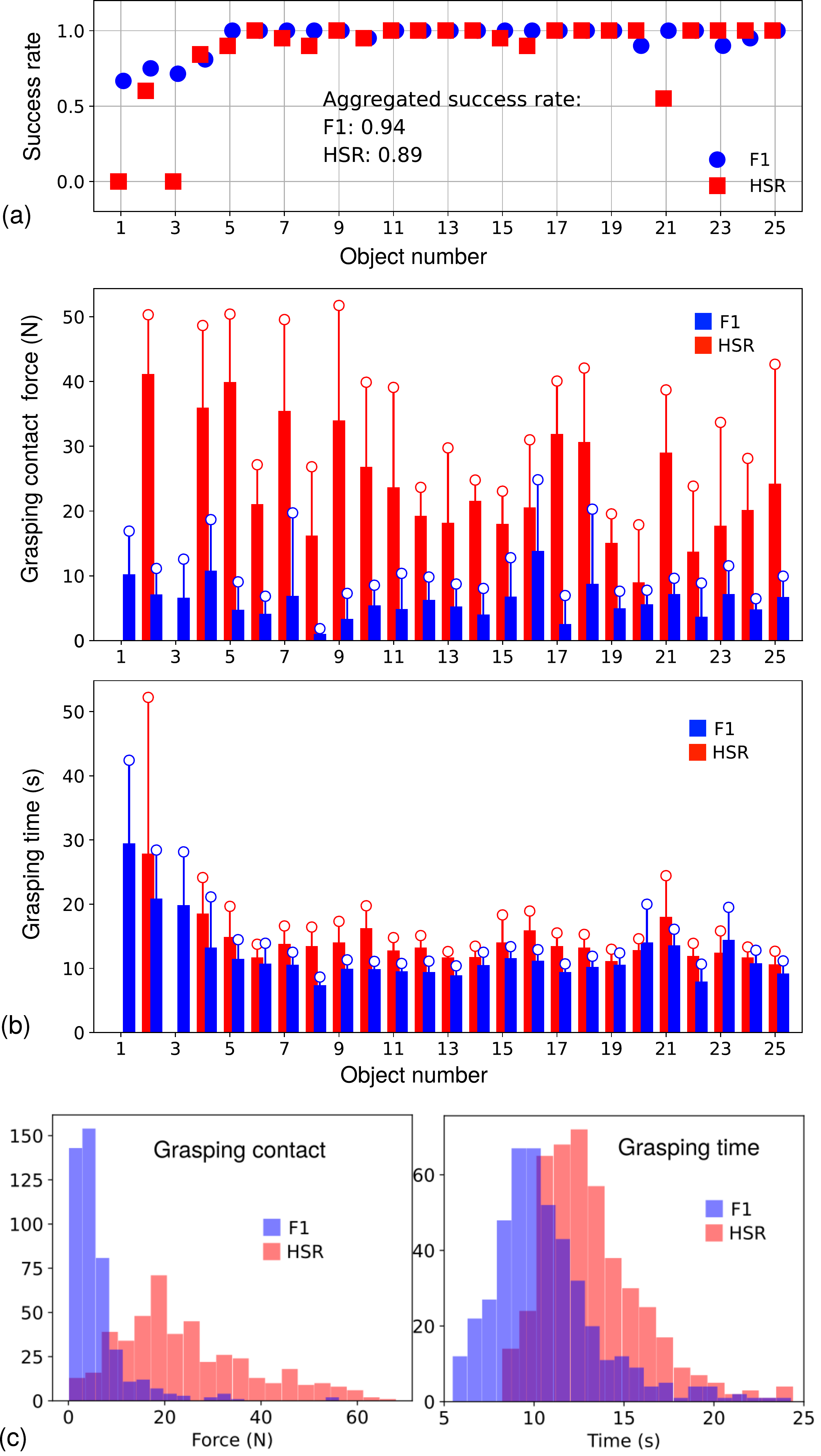}
		\vspace{-.002cm}
		\caption{
			Results of the teleoperation where each of the 25 objects were grasped between 20 to 21 times.
			(a) The aggregated success rates over all objects of the F1 and the original HSR gripper.
			(b) Individual peak contact force and time to grasp.
			(c) The histogram of the 460 grasps, excluding the objects \#1 and \#3 for which the HSR data is not available.
		}
		\label{fig:result_teleop}
	\end{figure}

	Figure \ref{fig:result_teleop}(a) summarizes the success rate individually. 
	The success rates of the original HSR gripper for the coin (\#1) and the washer (\#3) were zero.
	Despite our best efforts, the original HSR gripper could not appropriately grasp these two objects.
	In many trials, the torque controller to close the gripper would fail as the tips of the fingers would touch the surface before grasping the object, eventually leading the robot unresponsive as the fingers could not be closed and the gripper could not be retracted due to the ROS blocking service.
	The aggregated success rate including all objects was 89\% for the original HSR gripper and increased to 94\% when using the F1 hands.

	Figure \ref{fig:result_teleop}(b) shows the mean (bar height) and one standard deviation (circle) of the 20 trials in terms of the peak normal contact force and the grasping time while (c) shows the histogram over all trials and all objects (460 runs in each robot).
	The F1 hand showed lower contact force with the surfaces over all object cases. 
	The mode of the histogram of the F1 hand is 4.28 N while {\color{\FixColor}that of the original HSR gripper} is 18.78 N, representing a reduction of 77\% in peak force.
	Teleoperation time decreased in average by 3 seconds which makes grasping with the F1 hands 20\% faster.
	For the F1 hands, the time histogram has a mean $\pm$ one standard deviation value of 10.99 $\pm$ 4.06 seconds while {\color{\FixColor} that of the original HSR gripper} is  13.88 $\pm$ 5.83 seconds.
	
	{\color{\FixColor}
		We also examine the performance with the another expert operator, and found the results are quite consistent.
		The overall success rates, contact force, and grasping time averaged among the objects are 88\%, 21.43 N, and 13.49  $\pm$ 2.1 seconds for original HSR gripper and 95\%, 4.32 N, 11.2 $\pm$ 1.65 seconds, for F1 hand, respectively.}
	
	
	\begin{figure*}
		\centering
		\vspace{0.2cm}
		\includegraphics[width=0.99\linewidth]{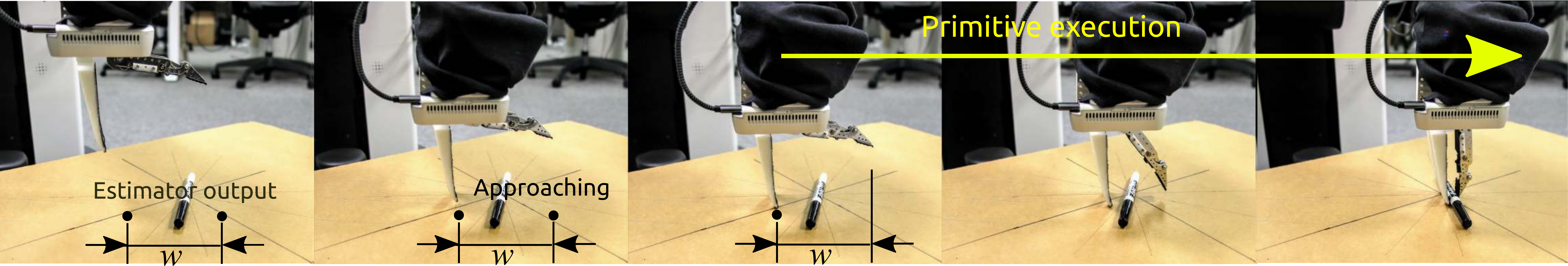}
		\vspace{-0.2cm}
		\caption{
			A sequence of snapshots of the F1 hand grasping a pen in autonomous mode. 
			The grasping pose was estimated with the GG-CNN.
			The entire action can be divided in two segments.
			In the first, the robot positions $O_{tip}$ as a pre-grasping motion.
			In the second, the robot executes the primitive control as it was described in Sec. \ref{sec:autonomous_grasping}.
		}
		\label{fig:snapshotgraspingpenggcnn}
	\end{figure*}
	
	{\color{\FixColor}
		Despite both robots being identical, the decrease in grasping time with the F1 hand was visible.
		To its advantage, the F1 hand does not require adjusting the height of the gripper while the fingers close, as is the case with the original HSR gripper, which saves time.
		Also, the primitive controller allows for a rough initial positioning of the hand as the adaptive movable finger assimilates the shape of the object while using the fixed finger as a supporting counter-part.
		As a result, the margin for relative error in the positioning of the F1 hand is larger than when grasping objects with the HSR gripper which leads to faster operations.
	}
	
	\subsection{Autonomous Grasping}
	
	Due to its popularity and simplicity, the GG-CNN~\cite{morrisonLearningRobustRealtime2020} was used to investigate the efficacy of the primitive described in Sec. \ref{sec:autonomous_grasping} in autonomous mode.
	We selected ten of the objects used in the teleoperation dataset. 
	We eliminated the small objects (\#1 to \#5) due to a lack of resolution in depth perception  of the RealSense D435i.
	For the bigger objects, we selected representative objects that lead to similar  pose estimation by the GG-CNN.
	As such, for the objects with a rim (\#8, \#12, \#17, \#22) we selected the \#17 as the most challenging one.
	For cylindrical objects (\#9, \#14, \#18, \#25) we used \#18  and \#25.
	For the two spherical objects (\#19, \#20) we selected \#20.
	We used the toy drill \#15 instead of the real one \#16.
	The airplane \#21 and the mustard \#23 were excluded due to their height because often the camera would be hit during the approaching phase or due to extraneous grasping poses\footnote{We acknowledge this problem can be solved by using a motion planner on the point cloud during pre-grasping and we leave such addition  as future work.}, which would require stopping the experiments, assessing, and eventually re-doing calibration procedures.

	Each object was placed individually and centered on a table in front of the robot and 20 trials were executed. 
	The initial object position was changed by rotating approximately 30$^\circ \sim$ 45$^\circ$ from the previous trial.
	The robot started by going to a home position where the camera on the wrist was used to capture the RGB-D image from a top-down configuration.
	Under this pose, the GG-CNN was used to infer the ($x,y,z$) and the $\theta$ parameter of the grasping pose.
	
\begin{figure}
		\centering
		\includegraphics[width=0.97\linewidth]{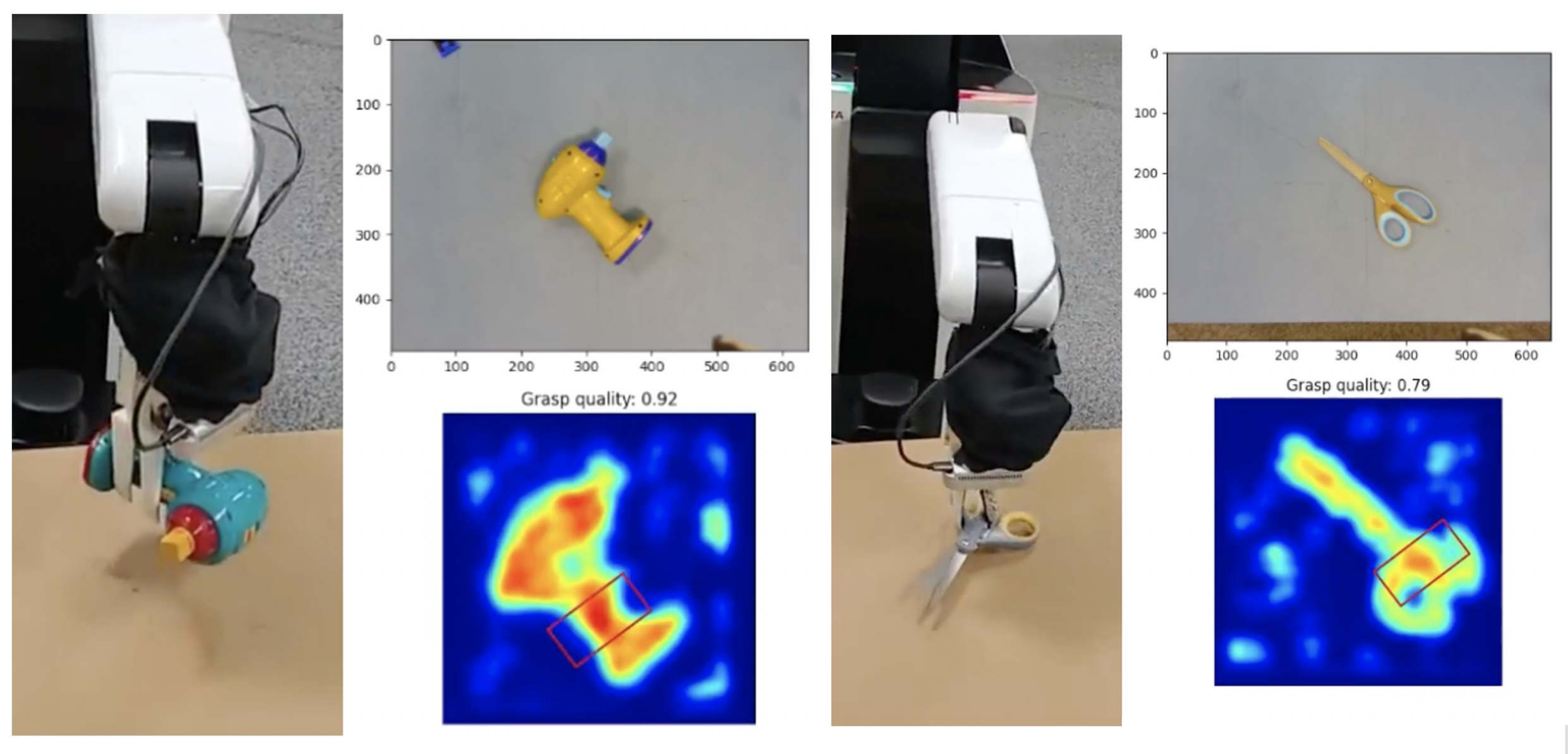}
		\vspace{-0.2cm}
		\caption{
			Experimental example of autonomous grasping using GGCNN. 
			The red {\color{\FixColor}rectangles} indicate the grasping points estimated by the GGCNN. 
			During operation, the robot first aligns the gripper orientation before starting the execution of the grasping. The adaptive mechanism increases the success rate of grasping even for hard-to-grasp objects such as the scissors on the right.
			Autonomous grasping experiments can be watched in the second part of the video here: \href{https://youtu.be/iWXXIX4Mkl8}{https://youtu.be/iWXXIX4Mkl8}
		}
		\color{black}
		\label{fig:testshot_ggcnn}
	\end{figure}

	Once the grasp was detected, the controller mapped the $(x,y,z, \theta)$  values to the grasping primitive shown in Fig. \ref{fig:controlstrategyposeestimator}.
	As shown in Fig. \ref{fig:snapshotgraspingpenggcnn}, the robot used a straight trajectory as a pre-grasping approach to move from its current position to the start of the primitive motion, before executing the grasping controller.
	Then, the robot attempted to lift the object 10 cm above the table (Fig. \ref{fig:testshot_ggcnn}).
	Human supervision was used to verify grasp success.
	To account for inaccuracies of the pose estimator and calibration errors, we used a wider version of the fixed-finger gripper (right finger in Fig. \ref{fig:robot_setup}(b)).

\begin{figure}
		\centering
		\includegraphics[width=0.97\linewidth]{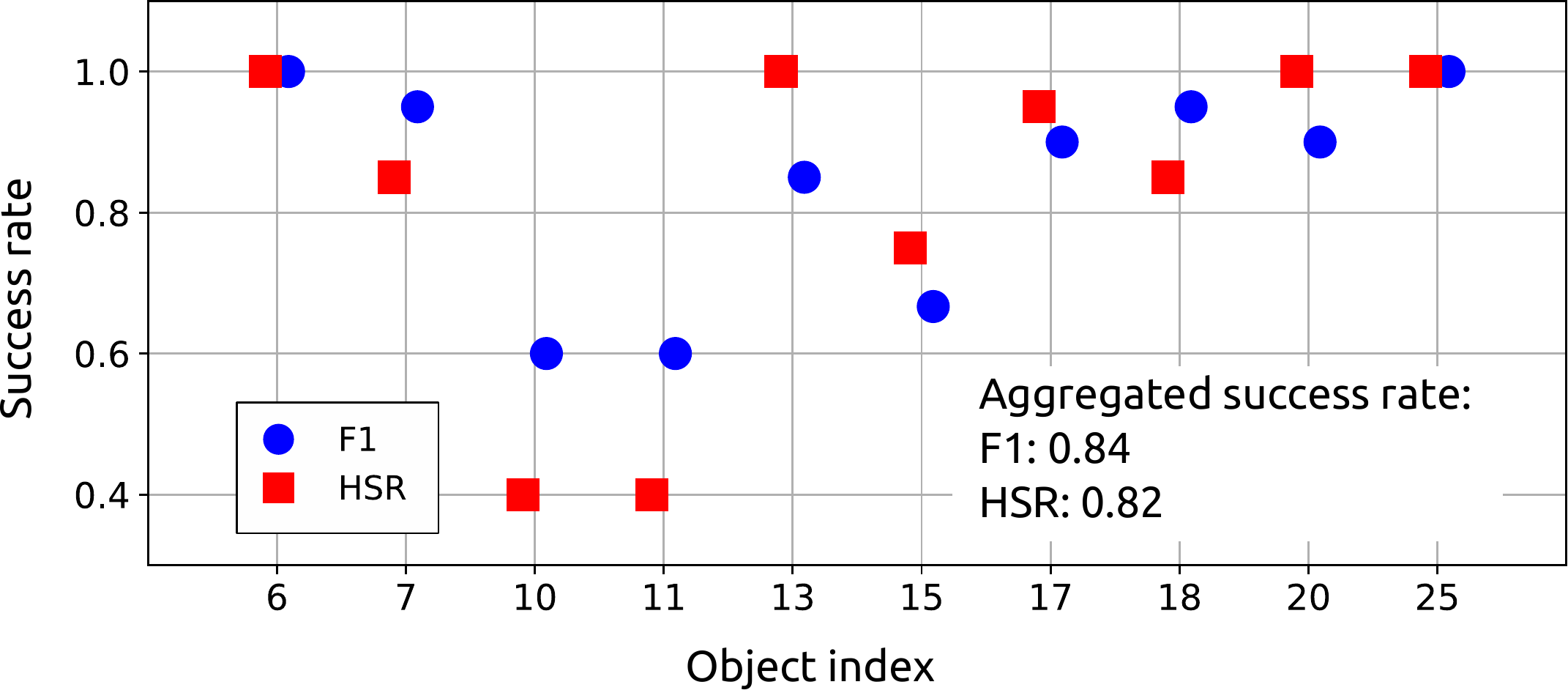}
		\vspace{-0.2cm}
		\caption{
			Success rates of autonomous grasping trials with both the original HSR gripper and the F1 hand.
			A similar trend was observed per object due to erratic and/or suboptimal grasp pose estimations from the GG-CNN which affected both grippers equally.
			The objects and their indices are indicated in Fig. \ref{fig:objects}.
		}
		\label{fig:successrate_ggcnn}
	\end{figure}

	The success rate with both hands shows no significant difference as indicated by Fig. \ref{fig:successrate_ggcnn} where the F1 hand and the HSR original gripper achieved an overall score of 0.84 and 0.82, respectively.
	Similar trends are observed on a per-object basis, an indication that the success disparity among objects is mainly due to the adequacy of the grasp pose estimation rather than the hand capabilities.
	Indeed, for certain object orientations, the estimator would estimate extraneous grasping orientations such as grasping a pen with the finger aligned with the pen's length, which led to failure regardless of the gripper in use.
	
	This experiment's result allows us to verify that our system could successfully use the output of a grasp pose estimator that assumes symmetric actuation by leveraging the whole-body motion of the robot with the use of the grasping primitive actions.
	The fact that similar success rates were achieved by the HSR original gripper, indicates that the  mapping by the controller did not decrease the success rate.

	\begin{figure*}
		\centering
		\vspace{0.2cm}
		\includegraphics[width=0.9850\linewidth]{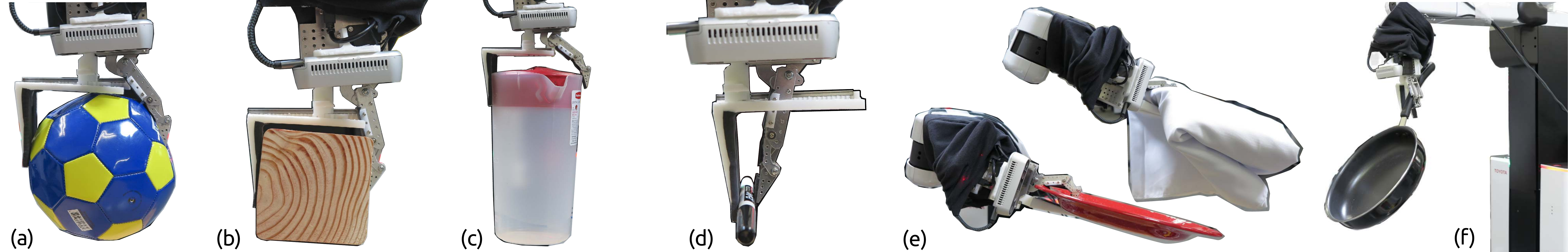}
		\vspace{-0.2cm}
		\caption{
			Challenging YCB objects. 
			(a-d) A fixed finger with a latch mechanism makes it possible to slide the fixed finger on a linear guide to increase the hand aperture. 
			(e) Grasping the plate and the table cloth are possible when enabling the roll motion of the hand.
			(f) Grasping the frying pan is still challenging due to the large momentum exerted on the wrist.
		}
		\label{fig:otherfingerdesigns}
	\end{figure*}

	\section{Extending Grasping to other YCB Objects} \label{sec:limitation_and_extension}
	
	We discuss some of the most challenging YCB objects not included in the experiments due to the current limitations of the system and how we plan to address them.
	Using the fingers of Sec. \ref{sec:experiments} the F1 hand was not capable of grasping some of the largest objects of the dataset. 
	Namely, the mini-soccer ball, the wooden block, and the jar.
	Currently, we are investigating a sliding fixed finger that uses a linear guide and a latch to allow for quick and easy change of the hand opening.
	This functionality increases the distance between fingers from 130.5 mm to 215 mm and makes it possible  to grasp not only bigger objects not previously possible---shown in Figs. \ref{fig:otherfingerdesigns}(a-c)---but also smaller objects such as a pen  as shown in (d).
	
	The plate and the table cloths are graspable but they require approaching the objects sideways, which was out of the scope of our top grasp study. Finally, the issue of the frying pan is having the position of its heavy center of mass (0.8 kg) far away from the handle and thus causing unstable grasps due to excessive momentum on the wrist.
	This could be solved with wide fixed finger designs that could act as a wide palm to provide support to the large momentum and remain as one of the challenges for our next iterations of hand design.

	\section{CONCLUSIONS}
	
	We introduced the F1 hand, a novel versatile two-fingered gripper with a fixed finger, and evaluated it as part of a robotic system. 
	This fixed finger facilitates the visual assessment of the relative position of the hand to the object during the grasping operation, thus enabling a more intuitive teleoperation.
	A controller was introduced to preserve the intuitive approach of centering objects between the fingers by mapping symmetry parameters to the asymmetric actuation of the hand.
	In total, together with teleoperation and autonomous graspings, approximately 700 grasps were executed with the F1 hands mounted on a Toyota HSR using the proposed control method.
	Compared to the original HSR gripper, under teleoperation, our hand provided faster grasping while generating significantly less force on the surfaces, with a slight increase in success rates. Also, when using an off-the-shelf grasp pose estimator, the autonomous grasping success rate was comparable to the original HSR gripper.

	\section*{ACKNOWLEDGMENT}
	
	The authors cordially express their gratitude to Mr. Shimpei Masuda, Dr. Tianyi Ko and Dr. Koji Terada in Preferred Networks, Inc. and Mr. Koichi Ikeda, Mr. Hiroshi Bito and Dr. Hideki Kajima in Toyota Motor Corporation for their assistance.
	
%

\begin{thebibliography}{10}
		\providecommand{\url}[1]{#1}
		\csname url@rmstyle\endcsname
		\providecommand{\newblock}{\relax}
		\providecommand{\bibinfo}[2]{#2}
		\providecommand\BIBentrySTDinterwordspacing{\spaceskip=0pt\relax}
		\providecommand\BIBentryALTinterwordstretchfactor{4}
		\providecommand\BIBentryALTinterwordspacing{\spaceskip=\fontdimen2\font plus
			\BIBentryALTinterwordstretchfactor\fontdimen3\font minus
			\fontdimen4\font\relax}
		\providecommand\BIBforeignlanguage[2]{{%
				\expandafter\ifx\csname l@#1\endcsname\relax
				\typeout{** WARNING: IEEEtran.bst: No hyphenation pattern has been}%
				\typeout{** loaded for the language `#1'. Using the pattern for}%
				\typeout{** the default language instead.}%
				\else
				\language=\csname l@#1\endcsname
				\fi
				#2}}
		
		\bibitem{grebensteinAntagonisticallyDrivenFinger2010}
		M.~Grebenstein, M.~Chalon, G.~Hirzinger, and R.~Siegwart, ``Antagonistically
		driven finger design for the anthropomorphic {{DLR}} hand arm system,'' in
		\emph{2010 10th {{IEEE-RAS International Conference}} on {{Humanoid
					Robots}}}.\hskip 1em plus 0.5em minus 0.4em\relax {IEEE}, 2010, pp. 609--616.
		
		\bibitem{akkayaSolvingRubikCube2019}
		I.~Akkaya, M.~Andrychowicz, M.~Chociej, M.~Litwin, B.~McGrew, A.~Petron,
		A.~Paino, M.~Plappert, G.~Powell, and R.~Ribas, ``Solving rubik's cube with a
		robot hand,'' \emph{arXiv preprint arXiv:1910.07113}, 2019.
		
		\bibitem{li_teleop_2017}
		Z.~Li, P.~Moran, Q.~Dong, R.~Shaw, and K.~Hauser, ``Development of a
		tele-nursing mobile manipulator for remote care-giving in quarantine areas,''
		in \emph{2017 IEEE International Conference on Robotics and Automation
			(ICRA)}.\hskip 1em plus 0.5em minus 0.4em\relax {IEEE}, 2017, pp. 3581--3586.
		
		\bibitem{maM2GripperExtending2016}
		R.~R. Ma, A.~Spiers, and A.~M. Dollar, ``M2 gripper: {{Extending}} the
		dexterity of a simple, underactuated gripper,'' in \emph{Advances in
			Reconfigurable Mechanisms and Robots {{II}}}.\hskip 1em plus 0.5em minus
		0.4em\relax {Springer}, 2016, pp. 795--805.
		
		\bibitem{achilliDesignSoftGrippers2020}
		G.~M. Achilli, M.~C. Valigi, G.~Salvietti, and M.~Malvezzi, ``Design of {{Soft
				Grippers}} with {{Modular Actuated Embedded Constraints}},'' \emph{Robotics},
		vol.~9, no.~4, p. 105, 2020.
		
		\bibitem{KoTendonDrivenPFN}
		T.~Ko, ``A tendon-driven robot gripper with passively switchable underactuated
		surface and its physics simulation based parameter optimization,'' \emph{IEEE
			Robotics and Automation Letters}, vol.~5, no.~4, pp. 5002--5009, 2020.
		
		\bibitem{bircherComplexManipulationSimple2021}
		W.~G. Bircher, A.~S. Morgan, and A.~M. Dollar, ``Complex manipulation with a
		simple robotic hand through contact breaking and caging,'' \emph{Science
			Robotics}, vol.~6, no.~54, p. eabd2666, May 2021.
		
		\bibitem{yaguchiDevelopmentAutonomousTomato2016}
		H.~Yaguchi, K.~Nagahama, T.~Hasegawa, and M.~Inaba, ``Development of an
		autonomous tomato harvesting robot with rotational plucking gripper,'' in
		\emph{2016 {{IEEE}}/{{RSJ International Conference}} on {{Intelligent
					Robots}} and {{Systems}} ({{IROS}})}.\hskip 1em plus 0.5em minus 0.4em\relax
		{IEEE}, 2016, pp. 652--657.
		
		\bibitem{teeple2020_hand}
		C.~B. Teeple, T.~N. Koutros, M.~A. Graule, and R.~J. Wood, ``Multi-segment soft
		robotic fingers enable robust precision grasping,'' \emph{The International
			Journal of Robotics Research}, vol.~39, no.~14, pp. 1647--1667, 2020.
		
		\bibitem{yoshimi2012_hand}
		T.~Yoshimi, N.~Iwata, M.~Mizukawa, and Y.~Ando, ``Picking up operation of thin
		objects by robot arm with two-fingered parallel soft gripper,'' \emph{IEEE
			International Workshop on Advanced Robotics and its Social Impacts(ARSO),},
		2012.
		
		\bibitem{watanabe2021_hand}
		T.~Watanabe, K.~Morino, Y.~Asama, S.~Nishitani, and R.~Toshima,
		``Variable-grasping-mode gripper with different finger structures for
		grasping small-sized items,'' \emph{IEEE Robotics and Automation Letter},
		vol.~6, no.~3, pp. 5673--5680, 2021.
		
		\bibitem{kobayashiDesignDevelopmentCompactly2019}
		A.~Kobayashi, J.~Kinugawa, S.~Arai, and K.~Kosuge, ``Design and {{Development}}
		of {{Compactly Folding Parallel Open-Close Gripper}} with {{Wide Stroke}},''
		in \emph{2019 {{IEEE}}/{{RSJ International Conference}} on {{Intelligent
					Robots}} and {{Systems}} ({{IROS}})}.\hskip 1em plus 0.5em minus 0.4em\relax
		{IEEE}, 2019, pp. 2408--2414.
		
		\bibitem{yamamoto2019development}
		T.~Yamamoto, K.~Terada, A.~Ochiai, F.~Saito, Y.~Asahara, and K.~Murase,
		``Development of {{Human Support Robot}} as the research platform of a
		domestic mobile manipulator,'' \emph{ROBOMECH journal}, vol.~6, no.~1, p.~4,
		2019.
		
		\bibitem{pastorUsing3dConvolutional2019}
		F.~Pastor, J.~M. Gandarias, A.~J. {Garc{\'i}a-Cerezo}, and J.~M.
		{G{\'o}mez-de-Gabriel}, ``Using 3d convolutional neural networks for tactile
		object recognition with robotic palpation,'' \emph{Sensors}, vol.~19, no.~24,
		p. 5356, 2019.
		
		\bibitem{maInhandManipulationPrimitives2016}
		R.~R. Ma and A.~M. Dollar, ``In-hand manipulation primitives for a minimal,
		underactuated gripper with active surfaces,'' in \emph{International {{Design
					Engineering Technical Conferences}} and {{Computers}} and {{Information}} in
			{{Engineering Conference}}}, vol. 50152.\hskip 1em plus 0.5em minus
		0.4em\relax {American Society of Mechanical Engineers}, 2016, p. V05AT07A072.
		
		\bibitem{zhangDoraPickerAutonomousPicking2016}
		H.~Zhang, P.~Long, D.~Zhou, Z.~Qian, Z.~Wang, W.~Wan, D.~Manocha, C.~Park,
		T.~Hu, and C.~Cao, ``{{DoraPicker}}: {{An}} autonomous picking system for
		general objects,'' in \emph{2016 {{IEEE International Conference}} on
			{{Automation Science}} and {{Engineering}} ({{CASE}})}.\hskip 1em plus 0.5em
		minus 0.4em\relax {IEEE}, 2016, pp. 721--726.
		
		\bibitem{fukaya2000_hand}
		N.~Fukaya, T.~Asfour, R.~Dillmann, and S.~Toyama, ``Design of the
		{{TUAT}}/{{Karlsruhe}} humanoid hand,'' \emph{IEEE/RSJ International
			Conference on Intelligent Robots and Systems (IROS)}, pp. 1754--1759, 2000.
		
		\bibitem{fukaya2013_hand}
		N.~Fukaya, T.~Asfour, R.~Dillmann, and S.~Toyama., ``Development of a
		five-finger dexterous hand without feedback control: The
		{{TUAT}}/{{Karlsruhe}} humanoid hand,'' \emph{IEEE/RSJ International
			Conference on Intelligent Robots and Systems (IROS)}, pp. 4533--4540, 2013.
		
		\bibitem{morrisonLearningRobustRealtime2020}
		D.~Morrison, P.~Corke, and J.~Leitner, ``Learning robust, real-time, reactive
		robotic grasping,'' \emph{The International Journal of Robotics Research},
		vol.~39, no. 2-3, pp. 183--201, Mar. 2020.
		
		\bibitem{mahlerDexnetDeepLearning2017}
		J.~Mahler, J.~Liang, S.~Niyaz, M.~Laskey, R.~Doan, X.~Liu, J.~A. Ojea, and
		K.~Goldberg, ``Dex-{N}et 2.0: {{Deep}} learning to plan robust grasps with
		synthetic point clouds and analytic grasp metrics,'' \emph{arXiv preprint
			arXiv:1703.09312}, 2017.
		
		\bibitem{murali20196}
		A.~Murali, A.~Mousavian, C.~Eppner, C.~Paxton, and D.~Fox, ``6-{{DOF}} grasping
		for target-driven object manipulation in clutter,'' in \emph{2020 IEEE
			International Conference on Robotics and Automation (ICRA)}, 2020, pp.
		6232--6238.
		
		\bibitem{breyerVolumetricGraspingNetwork2021}
		M.~Breyer, J.~J. Chung, L.~Ott, R.~Siegwart, and J.~Nieto, ``Volumetric
		{{Grasping Network}}: {{Real-time}} 6 {{DOF Grasp Detection}} in
		{{Clutter}},'' \emph{arXiv:2101.01132 [cs]}, Jan. 2021.
		
		\bibitem{calliYaleCMUBerkeleyDatasetRobotic2017a}
		B.~Calli, A.~Singh, J.~Bruce, A.~Walsman, K.~Konolige, S.~Srinivasa, P.~Abbeel,
		and A.~M. Dollar, ``Yale-{{CMU-Berkeley}} dataset for robotic manipulation
		research,'' \emph{The International Journal of Robotics Research}, vol.~36,
		no.~3, pp. 261--268, 2017.
		
	\end{thebibliography}
%

\end{document}